\documentclass[conference]{IEEEtran}
\usepackage[utf8]{inputenc}
\usepackage{cite}
\usepackage{amsmath,amssymb,amsfonts, amsthm}
\usepackage{mathtools}
\usepackage{algorithmic}
\usepackage{graphicx}
\usepackage{textcomp}
\usepackage[justification=centering]{caption}
\usepackage{subcaption}
\usepackage[ruled,vlined]{algorithm2e}
\usepackage{epsfig}
\usepackage{float}
\usepackage{color}
\usepackage{balance}
\usepackage{bbm}
\usepackage{textcomp}
\usepackage{enumitem}
\usepackage{url}
\usepackage{comment}
\usepackage[normalem]{ulem}
\usepackage[flushleft]{threeparttable}
\usepackage{footnote}
\usepackage{tikz}
\usetikzlibrary{patterns}
\usetikzlibrary{decorations, decorations.text,backgrounds}
\usepackage{dsfont}
\usepackage{algorithmic}
\usepackage{algorithm2e}
\floatstyle{ruled}
\newfloat{algorithm}{tbp}{loa}
\providecommand{\algorithmname}{Algorithm}
\floatname{algorithm}{\protect\algorithmname}

\newtheorem{proposition}{Proposition}
\newtheorem{definition}{Definition}

\newcommand\blfootnote[1]{%
  \begingroup
  \renewcommand\thefootnote{}\footnote{#1}%
  \addtocounter{footnote}{-1}%
  \endgroup
}

\newcommand{\fig}[1]{Fig. \ref{#1}}

\newcommand{\argmax}{\mathrm{arg\,max}}
\newcommand{\argmin}{\mathrm{arg\,min}}
\newcommand{\fnorm}[1]{\|{#1}\|_{\cal F}}
\newcommand{\fprod}[2]{\langle{#1, #2}\rangle_{\cal F}}

\newcommand{\mathrmbold}[1]{\boldsymbol{\mathrm{#1}}}

\newcommand{\ie}{\textit{i.e.~}}
\newcommand{\ien}{\textit{i.e.}}
\newcommand{\eg}{\textit{e.g.~}}

\newcommand{\norm}[1]{\lVert#1\rVert}

\def\dd{\mathrm{d}}

\def\rmB{\mathrm{B}}

\def\bfX{\mathbf{X}}
\def\bfY{\mathbf{Y}}

\def\diag{\mathrm{diag}}

\def\x{x}
\def\y{y}

\def\m{m}
\def\I{I}

\def\bbR{\mathbb{R}}
\def\bbC{\mathbb{C}}
\def\ballradius{r}
\def\calM{{\cal M}}
\def\cM{{\cal M}}
\def\calX{{\cal X}}
\def\calC{{\cal C}}

\def\frakB{\mathfrak{B}}
\def\calP{{\cal P}}

\def\calT{{\cal T}}

\def\dB{~\mathrm{dB}}

\title{Semantic Channel Equalizer: Modelling Language Mismatch in Multi-User Semantic Communications}

\author{\IEEEauthorblockN{Mohamed Sana, Emilio Calvanese Strinati}

\IEEEauthorblockA{CEA-Leti, Université Grenoble Alpes, F-38000 Grenoble, France\\
Email : \{mohamed.sana; emilio.calvanese-strinati\}@cea.fr}}

\newcommand{\titleheader}{This work has been accepted for publication in 2023 IEEE Global Communications Conference (GLOBECOM) SAC: Machine Learning for Communications}

\makeatletter
\def\ps@IEEEtitlepagestyle{%
\def\@oddhead{\mbox{}\scriptsize \titleheader \rightmark \hfil }%
}
\makeatother

\begin{document}
\maketitle\blfootnote{This work was partially funded by the French government under the France 2030 ANR program ``PEPR Networks of the Future" (ref. 22-PEFT-0010) and by the 6G-GOALS Project under the HORIZON program (no. 101139232).}  
\begin{abstract} 
We consider a multi-user semantic communications system in which agents (transmitters and receivers) interact through the exchange of semantic messages to convey meanings. In this context, languages are instrumental in structuring the construction and consolidation of knowledge, influencing conceptual representation and semantic extraction and interpretation. Yet, the crucial role of languages in semantic communications is often overlooked. When this is not the case, agent languages are assumed compatible and unambiguously interoperable, ignoring practical limitations that may arise due to language mismatching. This is the focus of this work. When agents use distinct languages, message interpretation is prone to semantic noise resulting from critical distortion introduced by \emph{semantic channels}. To address this problem, this paper proposes a new semantic channel equalizer to counteract and limit the critical ambiguity in message interpretation. Our proposed solution models the mismatch of languages with measurable transformations over semantic representation spaces. We achieve this using optimal transport theory, where we model such transformations as transportation maps. Then, to recover at the receiver the meaning intended by the teacher we operate \emph{semantic equalization} to compensate for the transformation introduced by the semantic channel, either before transmission and/or after the reception of semantic messages. We implement the proposed approach as an operation over a codebook of transformations specifically designed for successful communication. Numerical results show that the proposed semantic channel equalizer outperforms traditional approaches in terms of operational complexity and transmission accuracy.
\end{abstract}

\section{Introduction}
Wireless networked intelligence and services are fuelling the evolution towards sustainable smart society, industry, and economies. Their extraordinary success and operational effectiveness have led to game-changer innovations in communication technologies, computer science, AI and hardware, making intelligent cyber physical systems an essential support for our daily lives. Intelligence consumes large amounts of data which need critically scarce resources to be generated, transported through the network, verified, processed and memorized.
To accommodate such burgeoning new services, in parallel to the ongoing effort to investigate traditional approaches consisting in \eg exploring higher frequencies, semantic communication (SemCom) appears as an unlocking solution and is now envisioned as a potential key enabler of future 6G networks \cite{CalvaneseGOWSC2021}. This approach engineers communication systems to effectively recreate or infer the meaning of what has been communicated rather than to just optimize opaque data pipes aiming to reproduce exactly exchanged sequences of symbols. It can achieve significant source data compression gain, reducing the amount of data to be sensed, communicated, and/or processed, thereby, reducing energy consumption, and latency of wireless communication systems. Communication of meanings to enable effective cooperation between intelligent agents requires reliable exchanges of semantics \cite{SanaLearningSemantics}. This is possible thanks to a common or shared language (either natural or artificial), which endows messages (or observations of the world) with semantic meanings \cite{shi2021semantic}. 
Common language can be learned a priori \eg via a (supervised) learning procedure. It can also emerge from an interactive consensus based on \eg reinforcement learning (RL), by learning a communication protocol (an effective language to attain a goal driving cooperation) \cite{farshbafan2023curriculum}. This requires communication attempts, thus incurring errors and communication costs \cite{foersterAFW16a, ding2020learning}. Alternatively, common language can be suggested by the environment, where agents ground their language in representations of the observed world \cite{lin2021learning}. {Current work on semantic communications assumes that the interaction between agents is based on an agreed common language, possibly learned or given. This assumption considers that communication errors occur only because of \emph{i}) syntactic wireless channel, \emph{ii}) lossy semantic compression, or \emph{iii}) limited expressivity of semantic extraction and interpretation function, ruling how knowledge is extracted, represented and interpreted. However, when agents use distinct languages, our previous work \cite{SanaLearningSemantics} shows that communication is prone to semantic noise that might stimulate interpretation errors causing defective cooperation strategies \cite{wang2020mathematical}.} 

To the best of our knowledge, there is no existing framework that explicitly address the problem of language mismatch in semantic communications. It is the focus of this work, which proposes a new framework for semantic channel equalization (EQ). To do so, (\emph{1}) we first explore the implications of language mismatch (ruling knowledge representation and reasoning capabilities) and propose a new modelling of the semantic channel between cooperating agents. In our proposed modelling, a mismatch of languages translates to a mismatch of knowledge representation, captured by \emph{measurable transformations} introduced on semantic representation space. Hence, (\emph{2}) we propose a novel framework based on optimal transport (OT) theory, which models such transformations as transport plans that we learn to construct a codebook. With the learned codebook of transformations, compensating language mismatch becomes a problem of identifying and implementing the right transformation during communication. (\emph{3}) We achieve this by defining a new measure to quantify the risk of misinterpreting the conveyed meaning, which we optimize using a Bayes-optimal approach. {Eventually, this work provides a solid foundation for robust and effective multi-user semantic communications. It can also be applied to multi-task learning and knowledge transferability \cite{sana2021transferable}.}

\section{System model}
\begin{figure}[!h]
    \centering
    \includegraphics[width=\columnwidth]{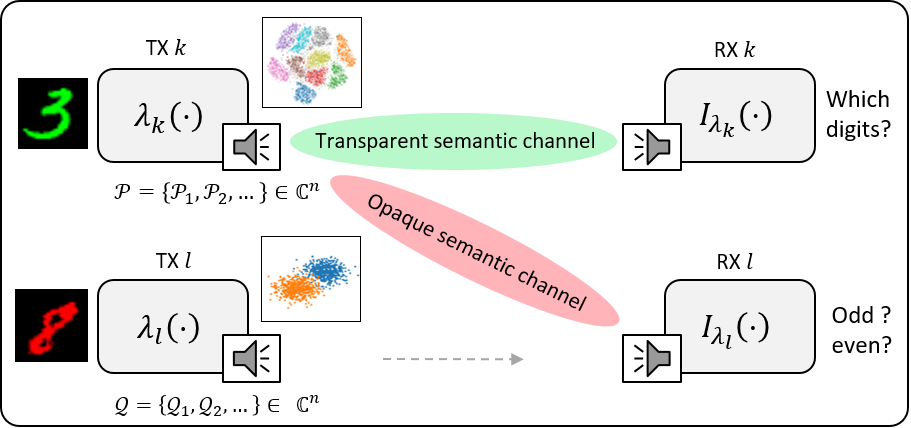}
    \caption{General system model.}
    \label{fig:sys-model}
\end{figure}
We consider the system model described in \fig{fig:sys-model}, where $K$ users employ distinct languages (different logic), here artificially modeled by neural networks (NNs), to endow messages (observations of the world) with semantic meanings. In this figure, two transmitters (TX) $k$ and $l$ observe images of digits and encode underlying information for two receivers (RX) $k$ and $l$, which try, respectively, to guess the digit on the image or its parity. Though the TXs act on the same observations of the world, they may structure differently information underlying observations for their respective RXs. Note that this reference example serves to easily explain our proposed method, which, obviously, is not limited to such use cases. 

\subsection{General framework: the central role of language}

We refer to the logic used by each TX to encode observations as its \emph{language generator}, the encoded information as \emph{semantic message}, and the mechanism used at the RX to recover the meaning of the information as \emph{language interpreter}. 

\noindent
\textbf{Language generator.} We model language generator as a mapping function $\lambda: \calM \rightarrow \calX$, which endows message or observation $\m\in\calM$ with semantic meaning $\x\in\calX$. Here, the message space $\calM$ and the semantics space $\calX$ are topological spaces, which can be continuous or discrete space. 
{Also, $\lambda$ can be deterministic or stochastic. In the latter case, we refer to $\lambda(\m)$ as a realization of this stochastic process (\eg a sampling from a generated probability distribution): $\lambda(\m)=\x\sim p(\x|m)$.} A language generator defines a way to partition a semantic space and associates to each partition a semantic meaning. Indeed, let $P=\{P_1, P_2, \dots\}$ be a partition of the probability space $(\calX, \frakB, \mu)$, where $\mu$ is a probability measure over $\calX$ and $\frakB$ is a $\sigma$-algebra on $\calX$ (\ie a nonempty collection of subsets of $\calX$ closed under complement, countable unions and intersections). 
We will often think of a partition of $\calX$ as a list of measurable sets that cover $\calX$. We refer to $P_i$ as the $i$-th atom of the partition $P$. Thus $\mu(P_i)$ is the probability measure of atom $P_i$. An atom may capture a certain level of semantics. For instance, for an image classification task, an atom may be related to the features (semantic symbols) associated to a certain class of an image.  
Different partitioning of the semantic space captures different level of semantics. Also, different languages may partition differently the semantic space.

\noindent
\textbf{Language interpreter.} Depending on the meaning the TX assigns to each transmitted data $\m$ (the atom to which $\m$ belongs to), generated semantic symbols get interpreted by the language interpreter using a mapping function, $\I_\lambda: \calX \rightarrow \cM_\lambda$. Note that the target message space $\cM_\lambda$ may differ from $\calM$. Thus, $\I_\lambda$ can but is not required to be the inverse function $\lambda^{-1}$. Besides, $\lambda^{-1}$ is not necessarily defined unless $\lambda$ is invertible: a bijective function, \ie a one-to-one correspondence mapping. However, $\lambda$ can be a one-to-many mapping: a message can be associated with different symbols, each conveying the same information. Conversely, $\lambda$ can be a many-to-one mapping: multiple messages are represented with the same symbol. 
Hence, $(\calM, \lambda, \I_\lambda,  \cM_\lambda, \calX, \frakB, \mu)$ fully defines a language, whose expressivity is crucial for effective communication. Indeed, language may be limited: either \emph{i}) not all messages of the world have a meaning in this language or \emph{ii}) can be efficiently represented by the associated language generator or \emph{iii}) the meaning accurately recovered by the language interpreter.
This work does not focus on learning language generators or interpreters, which we assume are given. Instead, we focus specifically on communication between intelligent agents (with reasoning and processing capabilities that follows a language logic) to convey semantic messages to each other. These messages are interpretable by the RX depending on the meaning the TX assigns to them. Correct interpretation is not necessarily guaranteed, unless the TX and RX agree on and implement a common communication strategy or plan. For instance, considering the aforementioned example, RX $k$ may correctly interpret semantic messages transmitted by TX $k$ as they use the same language. In this case, we say that the semantic channel between TX $k$ and RX $k$ is transparent: they agree on the same logic for encoding and interpreting information. In contrast, RX $l$ may fail to recover the meaning intended by TX $k$ (in communication between TX $k$ and RX $l$). In this case, the semantic channel between TX $k$ and RX $l$ is said to be opaque. It is the focus of this work, which proposes a new framework for semantic channel equalization. 

\begin{figure}
    \centering
    \includegraphics[width=\columnwidth]{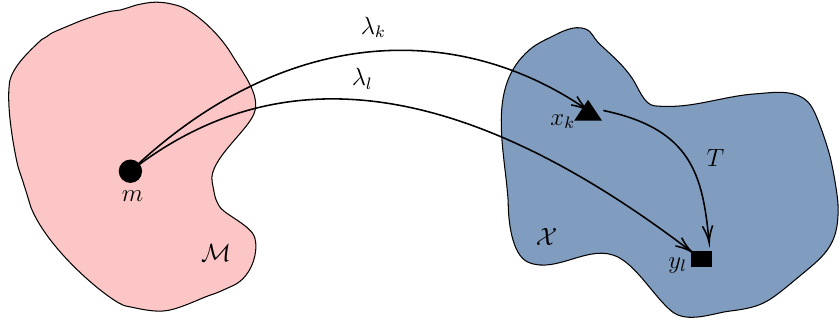}
    \caption{Illustration language mismatch between user $k$ and $l$.}
    \label{fig:language-mismatch}
\end{figure}

\subsection{Semantic channel modelling}

Consider communication between TX $k$ and RX $l$. Let $P=\{P_1, P_2,\dots\}$ and $Q=\{Q_1, Q_2,\dots\}$ be the two finite partitions of the probability space $(\calX, \frakB, \mu)$ defined by their language generators $\lambda_k$ and $\lambda_l$, respectively. 
Given a message $\m\in\calM$, TX $k$ encodes and transmits the associated meaning $\x_k = \lambda_k(\m)\in P_i$ to RX $l$, which interprets $\x_k$ in its own language as $\I_{\lambda_l}(\x_k)$. However, in RX $l$ language, data `$\m$' maps to $\y_l = \lambda_l(\m)\in Q_j$ (see \fig{fig:language-mismatch}). Thus, the atom $P_i$ in TX $k$ language is equivalent to atom $Q_j$ in RX $l$ language. As semantics are associated with atoms of partitions, RX $l$ may misinterpret received symbol $x_k$, especially when $x_k \not\in Q_j$. More specifically, when TX $k$ communicates with RX $l$, only the observations that get mapped into $P_i \cap Q_j$ can be correctly interpreted in RX $l$ language. Therefore, a mismatch of language translates to a mismatch of atoms of language partitions. 
Hereafter, we propose to model theses mismatches with \emph{measurable transformations}.

\begin{definition}
    Let $(\calX, \frakB)$ be a measurable space where $\frakB$ denotes a $\sigma$-algebra on $\calX$. A function $T: \calX\rightarrow\calX$ is said to be a measurable transformation if for every set $B\in\frakB$, \textbf{the preimage of $B$ under $T$ denoted $T^{-1}(B)$} is in $\frakB$; \ien:
    \begin{align}
        T^{-1}(B) \coloneqq \{\x\in \calX ~|~T(\x) \in B\}\in\frakB.
    \end{align}
\end{definition}

\begin{figure}
    \centering
    \includegraphics[width=0.75\columnwidth]{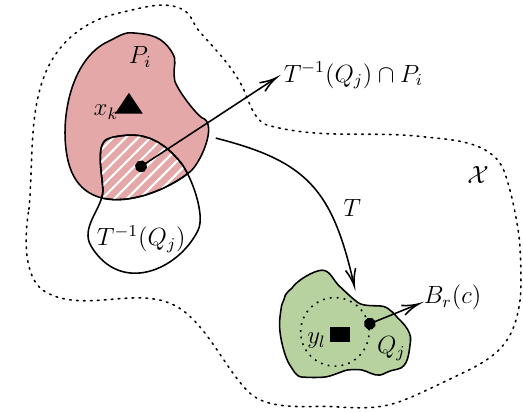}
    \caption{Partitioning of the semantic representation's space.}
    \label{fig:semantic-partitions}
\end{figure}

\begin{proposition}\label{prop:information-transfer} 
Let $P=\{P_1, P_2,\dots\}$ and $Q=\{Q_1, Q_2,\dots\}$ define two finite partitions of a probability space $(\calX, \frakB, \mu)$ such that $\mu(P_i)>0, ~\forall P_i\in P$. Let $T\in\calT$ be a measurable transformation. 
We define, $\forall ~(P_i,Q_j)\in P\times Q$:
\begin{align}\label{eq:info-transfer}
    \rho_{P_i\rightarrow Q_j}(T) = \frac{\mu(T^{-1}(Q_j)\cap P_i)}{\mu(P_i)}.
\end{align}
Intuitively, as show in \fig{fig:semantic-partitions}, $\rho_{P_i\rightarrow Q_j}(T)$ quantifies the fraction of the volume of $P_i$ that ends up in the set $Q_j$ (information transfer from $P_i$ to $Q_j$) under one iterate of $T$ \cite{Sinha19}. Since $\mu$ is a probability measure, we have $0\leq\mu(T^{-1}(Q_j)\cap P_i)\leq \mu(P_i)$, thus, $0\leq \rho_{P_i\rightarrow Q_j}(T) \leq 1$. To align atom $P_i$ with atom $Q_j$, it is sufficient to find a transformation $T$ that transports $x\in P_i$ into its equivalent semantic atom $T(x)\in Q_j$:  
\begin{align} \label{eq:info-optimization}
    \mathrmbold{P0}: \underset{T\in\calT}{\argmax}{\; \rho_{P_i\rightarrow Q_j}(T)}.
\end{align}
Transformation $T$ can be viewed as a language ``translator'' bridging the semantic gap between the source and target atom.
\end{proposition}

\begin{definition}[Language mismatch]
Let $J_P$ and $J_Q$ define two finite index sets of two language partitions $P=\{P_1, P_2,\dots\}$ and $Q=\{Q_1, Q_2,\dots\}$, respectively. Let $\kappa: J_P \rightarrow J_Q$ define a function, which associates each atom $i \in J_P$ of partition $P$ to the corresponding semantic atom $j\in\kappa(i) \subset J_Q$ in partition $Q$. Here, $\kappa(\cdot)$ can but is not required to be a one-to-one mapping. It can also be a many-to-one (\eg the digits $\{1, 3, 5, 7, 9\}$ are all odd numbers) or one-to-many mapping (\eg an even digit can be $\{0, 2, 4, 6, 8\}$). Let $T\in\calT$ be a measurable transformation. We define 
\begin{align}\label{eq:language-mismatch}
    \rho_0(T) = \sum_{i\in J_P} \sum_{j\in\kappa(i)} \rho_{P_i\rightarrow Q_j}(T).
\end{align} 
In particular, when $T$ is the identity function, $\rho_0$ measures the mismatch of source and target language partitions. 
\end{definition}
From Proposition \eqref{prop:information-transfer}, language mismatch can be captured by measurable transformations acting on $\calX$ (see \fig{fig:semantic-partitions}). These transformations model the semantic channel between users. To recover exactly at the RX the meaning intended by the TX sending a given symbol, either the RX manages to compensate for transformations introduced by the semantic channel (post-equalization), or the TX intentionally distorts intended meaning by selecting another symbol (pre-equalization). 

\section{Learning Measurable transformations}
In this section, we are interested in finding and describing measurable transformations for modelling the semantic channel between two users. Our approach relies on optimal transport (OT) theory to devise measurable transformations. We first introduce the original Monge-Kantorovitch problem, then its discrete formulation. Eventually, we propose a new optimization framework, which relaxes the original problem constraints to learn semantic transport maps.

\subsection{Background on the Monge-Kantorovitch Problem}
Let us consider a communication between two users employing distinct languages, referred to as the source and target language, each partitioning the semantic space $\calX$ into finite partitions $P=\{P_1, P_2,\dots\}$ and $Q=\{Q_1, Q_2,\dots\}$ respectively. Hereafter, we assume $\calX\subset \bbC^{n}$, where $n$ is the dimension of the \emph{complex} semantic space (Each transmitted data is represented with $n$ complex semantic symbols). For any atom $P_i$ of partition $P$ and $Q_j$ of partition $Q$, let $\mu_i \in \calP(P_i)$ and $\nu_j \in \calP(Q_j)$, where $\calP(A)$ is the set of all probability measures over $A$. For any message $\m$ mapped to $\x \in P_i$ by the source language, we seek a mapping $T: \calX \rightarrow \calX$, which transport $\x$ into the corresponding atom $Q_j$ in the target language, \ie such that $T(\x) \in Q_j$. Specifically, we seek a  transport map $T$ that pushes $\mu_i$ onto $\nu_j$, \ie $T_\#\mu_i = \nu_j$, where $T_\#\mu_i$, is the \emph{image measure} of $\mu_i$ by $T$, defined such that $T_\#\mu_i(A) = \mu_i(T^{-1}(A)), ~\forall A \subset \calX$. Let $d: \calX \times \calX \rightarrow \bbR^+$ be a metric (\eg Euclidean distance). 
The optimal transport map, when it exists, can be found by solving the following Monge transportation problem:
\begin{align}
    \underset{T}{\argmin} \; \int_{P_i} d(\x, T(\x)) \dd\mu_i(\x), ~\mathrm{s.t.} ~ T_\#\mu_i = \nu_j.
\end{align}
The existence of such map is not always guaranteed as when for \eg $\mu_i$ is supported on Dirac distributions and $\nu_j$ is not.

\noindent
\textbf{Kantorovitch relaxation.} The Kantorovitch problem seeks a general coupling $\gamma\in\Gamma_{i,j}$ between $P_i$ and $Q_j$ by solving:
\begin{align}
    \underset{\gamma\in\Gamma_{i,j}}{\argmin}{\; \int_{P_i\times Q_j} d(\x, \y)} \dd \gamma(\x, \y).
\end{align}
Here, $\Gamma_{i,j}$ is the set of all probabilistic measures in $\calP(P_i \times Q_j)$ with marginals $\mu_i$ and $\nu_j$. In contrast to $T$, the optimal coupling $\gamma$ exists but may not be unique \cite{villani2009optimal}.

\noindent
\textbf{Case of discrete distributions.} Usually the exact distribution $\mu_i$ or $\nu_j$ is not available. Only collections of samples from those distributions are accessible. Let $\bfX_i=\{\x_{i,k} \in P_i,~k=1,\dots, N_\bfX\} \sim \mu_i$ and $\bfY_j=\{\y_{j,k} \in Q_j,~k=1,\dots, N_\bfY\} \sim \nu_j$ define two finite samples from atom $P_i$ and $Q_j$. The corresponding empirical distribution can be written as $\mu_i = \sum_{k=1}^{N_{\bfX}} p_k \delta_{\x_{i,k}}$ and $\nu_j = \sum_{k=1}^{N_{\bfY}} q_k \delta_{\y_{j,k}}$,
where $\delta_{\x}$ denotes a unit point mass (Dirac) at the point $x\in\bbC^n$; $p_k$ and $q_k$ are probability masses associated to the $k$-th sample in $\bfX_i$ and $\bfY_j$, respectively, and belong to the probability simplex. 
The set of probabilistic coupling between those two distributions is the set of doubly stochastic matrices $\Gamma_{i,j}$ defined as:
\begin{align}
    \hat{\Gamma}_{i,j} = \{\gamma \in (\bbR^+)^{N_{\bfX}\times N_{\bfY}} ~|~ \gamma\mathds{1}_{N_{\bfY}}=\mu_i, \gamma^T\mathds{1}_{N_{\bfX}}=\nu_j\},
\end{align}
where $\mathds{1}_{n}$ is a $n$-dimensional vector of ones. The Kantorovitch formulation of the optimal transport reads:
\begin{align}
    \underset{\gamma\in\hat{\Gamma}_{i,j}}{\argmin}{\; \fprod{\gamma}{\mathrmbold{D}} },
\end{align}
where $\fprod{\cdot}{\cdot}$ denote the Frobenius dot product and $\mathrmbold{D}\in(\bbR^+)^{N_{\bfX}\times N_{\bfY}}$ is the cost matrix related to the metric $d(\cdot,\cdot)$.

\subsection{Learning semantic transport maps}

Once $\gamma$ is obtained, the source samples $\bfX_i$ can then be transported into the target distribution using coupling-dependent barycentric mapping $T_\gamma$, which transport source sample $\bfX_i$ into the convex hull of target sample $\bfY_j$. 
When $d$ is the squared $\ell_2$ distance \ie, $d(x,y) = \norm{x-y}_2^2$ and the samples are drawn i.i.d from $\mu_i$ and $\nu_j$, which we now assume, the mapping can be written in the following matrix form \cite{perrot2016mapping}:
\begin{align}
    T_\gamma(\bfX_i) = \diag(\gamma\mathds{1}_{N_{\bfY}})^{-1} \gamma \bfY_j = N_{\bfX} \gamma \bfY_j.
\end{align}
Hence, the mapping $T_\gamma$ depends on the coupling, which needs to be computed for every new data sample to be transported. To address this problem, similar to \cite{perrot2016mapping}, we propose to jointly learn the probabilistic coupling $\gamma\in\hat{\Gamma}_{i,j}$ and a transformation $T\in\calT$, coupling-free, which approximates the barycentric mapping $T_{\gamma}$.
However, in our proposed language modelling, semantics are associated to the atoms of language partition. While the source language maps data $m\in\calM$ into its semantic representation $\x\in P_i$, the target language language maps it into $\y \in Q_j$. Thus, what matters is to find a transformation $T$, which correctly transports $x$ into the equivalent semantic atom, $T(x) \in Q_j$, where it can be correctly interpreted by the target language. Perfect alignment through the transformation $T$ of (the points of) these two atoms is not needed: there is no need to guarantee $\y = T(\x)$ but rather that $\rho_{P_i\rightarrow Q_j}(T) = 1$. Therefore, we propose to solve the following problem:
\begin{align} \label{eq:new-transport-optimization}
    \mathrmbold{P1}: \underset{\gamma\in\hat{\Gamma}_{i,j}, T\in\calT}{\argmin}{\; f_{i,j}^{(\ballradius)}(\gamma, T)} + \beta h(T),
\end{align}
where $\fnorm{\cdot}$ and  denote the Frobenius norm; $\alpha, ~\beta > 0$ are some hyper-parameters that trade-off the optimization; $h(T)$ is a regularization term to ensure a better generalization of $T$. Here, in contrast to \cite{perrot2016mapping}, we define $f_{i,j}^{(\ballradius)}(\gamma, T) = \fnorm{T(\bfX_i) - {N_{\bfX} \gamma \rmB_\ballradius(\bfY_j)}}^2 + \alpha\fprod{\gamma}{\mathrmbold{D}}$, where ${\rmB_\ballradius(\bfY_j)}$ denotes a \emph{barycentric mapping} of the samples in $\bfY_j$ into a ball of center $c$ and radius $\ballradius$, contained in the convex hull of $\bfY_j$, as shown in \fig{fig:semantic-partitions}. The center $c$ of this ball can but is not required to be the centroid of $\bfY_j$, which we adopt in the following. Thus, in contrast to \cite{perrot2016mapping}, we seek a transport map $T$ that transports samples $\bfX_i$ from atom $P_i$ of partition $P$ into a ball ${\rmB_\ballradius(\bfY_j)}$ of the target sample $\bfY_j$\footnote{{This is equivalent to defining a new metric $d(x,y)= \norm{x-\rmB_\ballradius(y)}_2^2$}}. The radius of this ball controls the confidence at the receiver in interpreting the meaning intended by the sender. It also robustifies transmitted semantic symbols against syntactic channel noise. {In the following, we define $\rmB_\ballradius(\bfY_j) = c + \ballradius ( \bfY_j - c)$, where $c=\mathbb{E}_{y\sim\nu_j}[\bfY_j]$. In particular, $\rmB_1(\bfY_j) = \bfY_j$, in which case, Eq. \eqref{eq:new-transport-optimization} falls back to the original formulation in \cite{perrot2016mapping}. Also we consider only linear transformations: $\mathcal{T} = \{T, \exists \mathbf{A}\in\bbC^{n\times n}, \mathbf{b}\in \bbC^n, \forall \x\in\calX,~T(\x) = \mathbf{A}\x+\mathbf{b}\}$. We show that non-linear language mismatch can be compensated by operating a low-complexity codebook of biased linear transformations.}

\section{Semantic channel EQ using a codebook}

\subsection{Design of a codebook of transformations}
 
Let $P=\{P_1, P_2,\dots\}$ and $Q=\{Q_1, Q_2,\dots\}$ define two language partitions. We now assume that $\kappa(\cdot)$ defined in \eqref{eq:language-mismatch} is either a one-to-one or a many-to-one mapping. In this case, $\forall i\in J_P$, $\kappa(i)$ is a singleton, allowing us to denote with $Q_{\kappa(i)}$ the equivalent atom in the target language (target atom) of the source atom $P_i$. Then, we construct a codebook of transformations 
\begin{align}\label{eq:codebook-construction}
\calC_{\calT}= \left\{T_i= \mathrm{solve}\;(\mathrmbold{P1}),~\forall (i, \kappa(i)), ~i\in J_P\right\},
\end{align}
obtained by solving Problem \eqref{eq:new-transport-optimization} for each couple $(i,\kappa(i))$ of source and target atom. 
Thus, the size of the codebook coincides with $N_P$, the number of semantic atoms of the source language, and does not depend on $N_Q$. {To ease the notation, in the following, we write $\rho_{i}(T_k) = \rho_{P_i\rightarrow Q_{\kappa(i)}}(T_k), ~\forall T_k\in \calC_{\calT}$, and, we define $\boldsymbol{\rho} = (\rho_{i}(T_k))_{i ,k \in J_P}$, referred to as the \emph{information transfer matrix}, which captures the codebook efficiency. Eventually, we define the entropy of the codebook as:} 
\begin{align}
    H_\mu(\calC_{\calT}) &= -\frac{1}{N_P} \sum_{T\in \calC_{\calT}}  \rho_{i}(T)\log(\rho_{i}(T))\\ \nonumber
    &+(1-\rho_{i}(T))\log(1-\rho_{i}(T)).
\end{align}

\subsection{Bayes-optimal selection of Codebook transformations}
We focus on semantic channel pre-equalization. In this case, let $m\in \calM$ be the observation that the sender $k$ wants to convey to the receiver $l$. Instead of transmitting $x=\lambda_k(m)$, the sender intentionally distorts meaning intended at the receiver by transmitting another semantic symbol $y=T(x)$ obtained by appropriately selecting a transformation $T\in\calC_{\calT}$. Now, we are interested in a probabilistic selection strategy $\pi \in \Pi$, from a class of parametric probabilistic functions $\Pi$, such that $\pi(m, T)$ specifies the probability for selecting 
a transformation $T\in\calC_{\calT}$ given data $m\in \calM$. To do so, we define 
\begin{align}
    R(\pi, m) &= 1 - \mathbb{E}_{\substack{T\sim\pi(m, T)}}\left[ \sum_{i\in J_P} \mu(P_i|m) \cdot \rho_{i}(T)\right], \label{eq:risk}
\end{align}
where $\mu(P_i|m)$ is the probability that the semantic symbol $\x=\lambda_k(\m)$ associated to data $m$ belongs to atom $P_i$ of the sender language partition $P$. Here, $R(\pi, m)$ 
quantifies the \emph{a priori} (before syntactic channel) risk of misinterpretation at the RX by the target language when we employ strategy $\pi$ for transmitting data $m$.
In particular, $R(\pi, m) \in [0,1]$, where $R(\pi, m)=0$ indicates zero risk of misinterpretation at the RX when strategy $\pi$ is employed for transmitting $\m$ (as opposed to $R(\pi, m)=1$).
Then, we propose the following Bayes optimization to minimize the average risk of misinterpretation:
\begin{align}\label{eq:bayes-opt}
    \mathrmbold{P2}: \underset{\pi\in\Pi}{\argmin}{\; \mathbb{E}_{\substack{m\sim p_{\calM}(m)}}[R(\pi,m)]},
\end{align}
where $p_{\calM}(m)$ is the probability that the source $\calM_k$ emits data $m$. Problem \eqref{eq:bayes-opt} can be efficiently and optimally solved by opportunistically maximizing $R(\pi, m)$ for each transmitted data $\m$ using Algorithm \ref{alg:risk} to obtain a deterministic policy. Future work will explore stochastic transformation selection, by learning a unified parametric policy using \eg RL. 

\begin{algorithm}[!h]
\SetAlgoLined
\textbf{S1.} Initialize the codebook $\calC_{\calT}=\{T_i,~\forall i~J_P\}$.\\
\textbf{S2.} Observe new data $m$ and let $\mathrmbold{u} = [\mu(P_i|m), ~\forall i\in J_P]^T$.\\
\textbf{S3.} Compute $i^*={\argmax}_{i\in J_P}{\; \boldsymbol{\rho}^T \mathrmbold{u}}$.\\
\textbf{S4.} Select and apply $T_{i^*}$.
\caption{Transformation selection}
\label{alg:risk}
\end{algorithm}
\vspace{-0.4cm}

\section{Numerical results} 
We adopt the colour-MNIST dataset \cite{foersterAFW16a}, consisting of images of digits of size $3\times28\times28$ randomly colored red, green, blue, and yellow. In this context, we consider our reference scenario of \fig{fig:sys-model} where two TXs $k$ and $l$ observe the pixels value of random image, while the colour label and digit value are hidden, and encode underlying information for their respective RXs $k$ and $l$, which try to infer the digit on the image or its parity. Accordingly, $\lambda_k$ partitions the semantic space $\mathbb{C}^n$ into $N_P=10$ atoms corresponding to the $10$ digits and $\lambda_l$ into $N_Q=2$ atoms. 
While images colour information is not used in interpretation, it introduces semantic noise, bringing additional complexity in semantic symbols representation. 
We use $N_{\bfY}=5N_{\bfX}=5000$ image samples and set $r=1$ (default), $n=2$, thus, encoding each colour image with only two complex semantic symbols. \emph{We assume language generators and interpreters are given, modeled as auto-encoders, and obtained after an end-to-end learning process, minimizing a cross-entropy loss as in \cite{SanaLearningSemantics}}. Language generators are composed of a convolutional layer of dimension $3n$, followed by a rectifier linear unit and a max-pool operator, another convolutional layer with dimension $8n$, followed by another max pooling operator, a dropout layer with probability $0.5$, and a rectifier linear unit layer, whose output is flattenned to feed a final linear layer of dimension $2n$. In contrast, each language interpreter only includes one linear layer of dimension $2n$. In addition, all convolutional layers use a kernel size equal to $5$, and the max pool operators use a $2$-sized kernel and stride. The power of the complex symbols $\x$ outputted by language generators is normalized so that $\mathbb{E}[\x]=1$. Once trained, the accuracy of the communication between TX $k$ and RX $k$ under error-free syntactic channel is $A_k=78.4\%$, whereas for TX $l$ and RX $l$ it is $A_l=94.4\%$. The $100\%$ accuracy is not attained in particular due to the noise in data and the limited expressivity of the language generator and interpreter. Now, we are interested in the communication between TX $k$ and RX $l$, grounded to semantic errors, as they are not trained to match each other logic: their semantic channel is thus opaque, requiring equalization. 
\begin{figure}[!t]
    \centering
    \includegraphics[width=0.63\columnwidth]{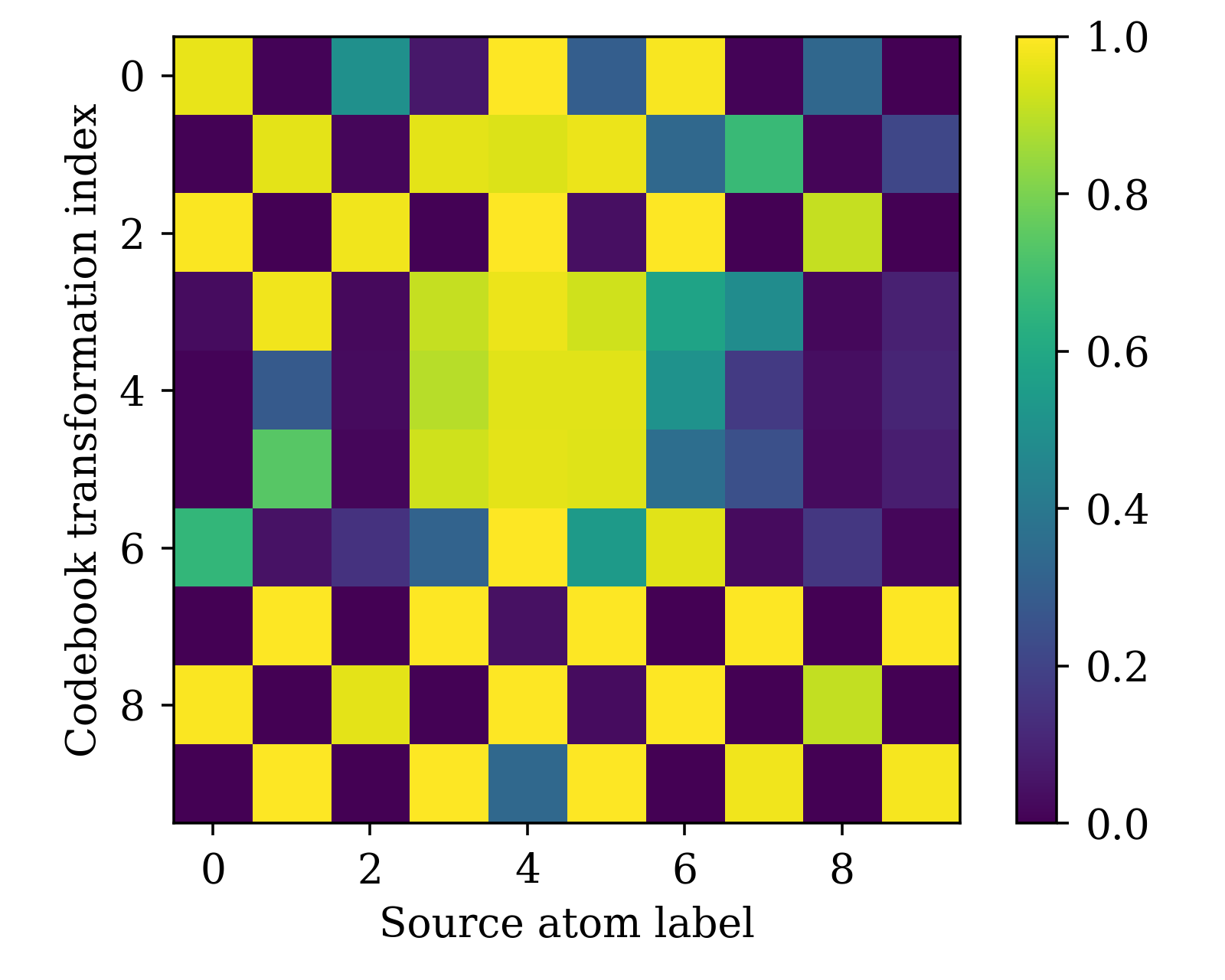}
    \caption{Information transfer matrix $\boldsymbol{\rho}$.}
    \label{fig:information_transfer}
\end{figure}

\vspace{0.1cm}
\noindent
\textbf{Operational zone of the codebook.} We assess the operational zone of the codebook constructed by solving Eq. \eqref{eq:codebook-construction} using the Algorithm proposed in \cite{perrot2016mapping} by setting $\alpha=\frac{0.1 n N_{\bfX}}{\max(\mathrmbold{D})}$, $\beta={10^{-8} n^{-1} N_{\bfX}}$, and $h(T) = \fnorm{\mathrmbold{A} - \mathrmbold{I}}^2$, where $\mathrmbold{I}$ is the identity matrix of size $n$. Then, we compute the associated information transfer matrix $\boldsymbol{\rho}$ shown on \fig{fig:information_transfer}. In practice, this is done \emph{offline} using a Monte-Carlo sampling. 
From \fig{fig:information_transfer}, the yellow diagonal in the matrix indicates that the optimized transformations in the codebook are successful in transporting samples from source atoms to target atoms.. \fig{fig:information_transfer} also suggests that there exist equivalent transformations in the codebook. In other words, for a given source atom $i$ and its associated target atom $\kappa(i)$, there exists more than one transformation with equal probability for transporting samples from $P_i$ into $Q_{\kappa(i)}$. For instance, for the atom with label $0$, the transformation at index $0$, $2$, and $8$ exhibits similar performance. Similarly, there exist orthogonal transformations. For instance, the transformations at index $1$, $3$, $4$, $5$, $7$, and $9$ cannot accurately transport the sample from the source atom $0$, to the associated target atom. These latter results suggest that the complexity can be further reduced by efficiently pruning transformations, which we will address in future work.

\begin{figure}[!t]
    \centering
    \includegraphics[width=0.98\columnwidth]{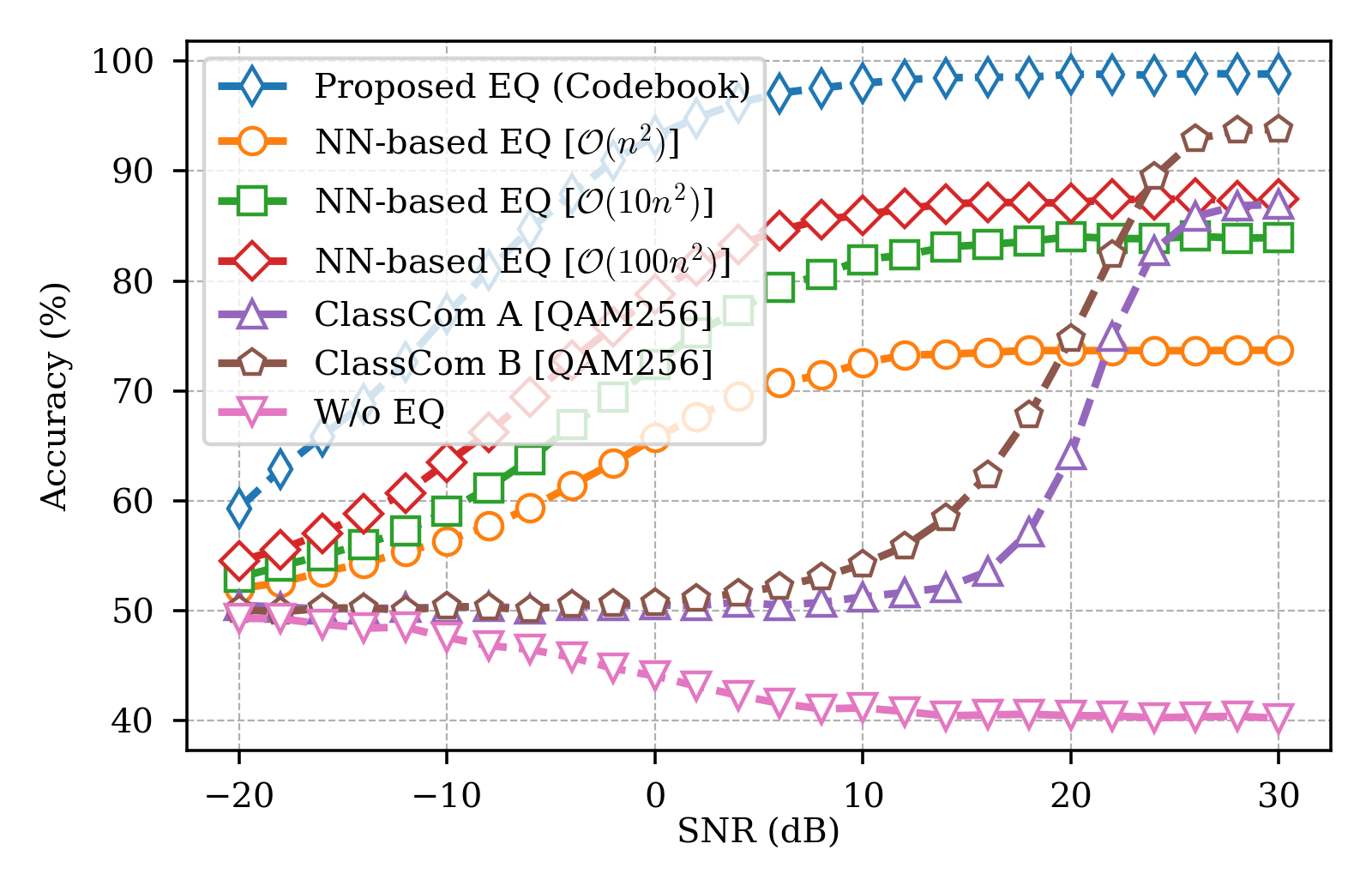}
    \caption{Performance comparison of equalization techniques.}
    \label{fig:perf_comparison_eq}
\end{figure}

\vspace{0.1cm}
\noindent
\textbf{The need for compensating language mismatch.} To show the effectiveness of our proposed equalization method, we compare it to different benchmarked approaches (see \fig{fig:perf_comparison_eq}):

\textbf{\emph{ClassCom}}: In the classical approach, TX $k$ sends binary encoded images to RX $l$, which infers the parity information from the received noisy images. In method \textbf{A}, the RX first infers the digit by applying $\I_{\lambda_k}\circ\lambda_k(\cdot)$ and then decodes its parity, whereas in method \textbf{B} it directly infers the parity by applying $\I_{\lambda_l}\circ\lambda_l(\cdot)$. This approach employs quadrature amplitude modulation (QAM) of order $256$, resulting in approximately $4.6 n$ symbols per transmitted image. Obviously, method \textbf{B} exhibits better performance than \textbf{A}, as it does not require decoding the digit first. However, the two classical approaches are strongly impacted by noise in the syntactic channel. At $0\dB$ SNR, our proposed solution reaches $90\%$ accuracy compared to $50\%$ accuracy for \emph{ClassCom} methods, which corresponds to random guessing of the digit parity. Even under error-free syntactic channel, the accuracy of method \textbf{B} caps to $A_l=94.4\%$ whereas for method \textbf{A}, it is $88\%$, corresponding $\approx A_k + \frac{4}{9}(1-A_k)$. Indeed, if the RX misinterpreted the digit, the probability that it misinterprets the parity is $\frac{5}{9}$. This result further confirms the value of semantic-based communication, which requires, however, semantic equalization mechanisms. %

\textbf{\emph{SemCom without EQ}}: The semantic channel is not equalized; that is, the codebook contains only the identity transformation. In this context, \fig{fig:perf_comparison_eq} shows that under an error-free syntactic channel, a communication between TX $k$ and RX $l$ results in $40.4\%$ accuracy. This performance is even below the $50\%$ accuracy expected for a random guess of the digit parity by RX $l$ from a uniform distribution of symbols in the received constellation, which can be obtained by noising the syntactic channel. This poor accuracy is a result of the mismatch of semantic representation space and motivates the need for semantic channel equalization \cite{SanaLearningSemantics}.

\textbf{\emph{SemCom with NN-based EQ}}: The codebook contains only one transformation, modeled as an NN comprising one linear layer of dimension $2n$, thus having substantially the same complexity (expressed as the number of parameters) as our proposed linear transformations \ie $\mathcal{O}(n^2)$. We train the NN by minimizing the cross-entropy loss, in an end-to-end fashion using the language generator $\lambda_k$ together with the language interpreter $\I_{\lambda_l}$. During the training, the weights of the language generator and interpreter are frozen so that they are not impacted by the learning procedure. \fig{fig:perf_comparison_eq} shows that, at $0\dB$ SNR, our proposed solution exhibits $30\%$ accuracy improvement compared to the NN-based equalizer. Also, we evaluate the case where the NN-based equalizer is $\tau$-times more complex (\ie has $\tau$-times more weights). In this case, it comprises two feed-forward layers of dimension $\tau n$ and $2n$, respectively. Even when the complexity of the NN increases, the accuracy of the communication does not exceed $88\%$.

\begin{figure}[!t]
    \centering
    \includegraphics[width=0.98\columnwidth]{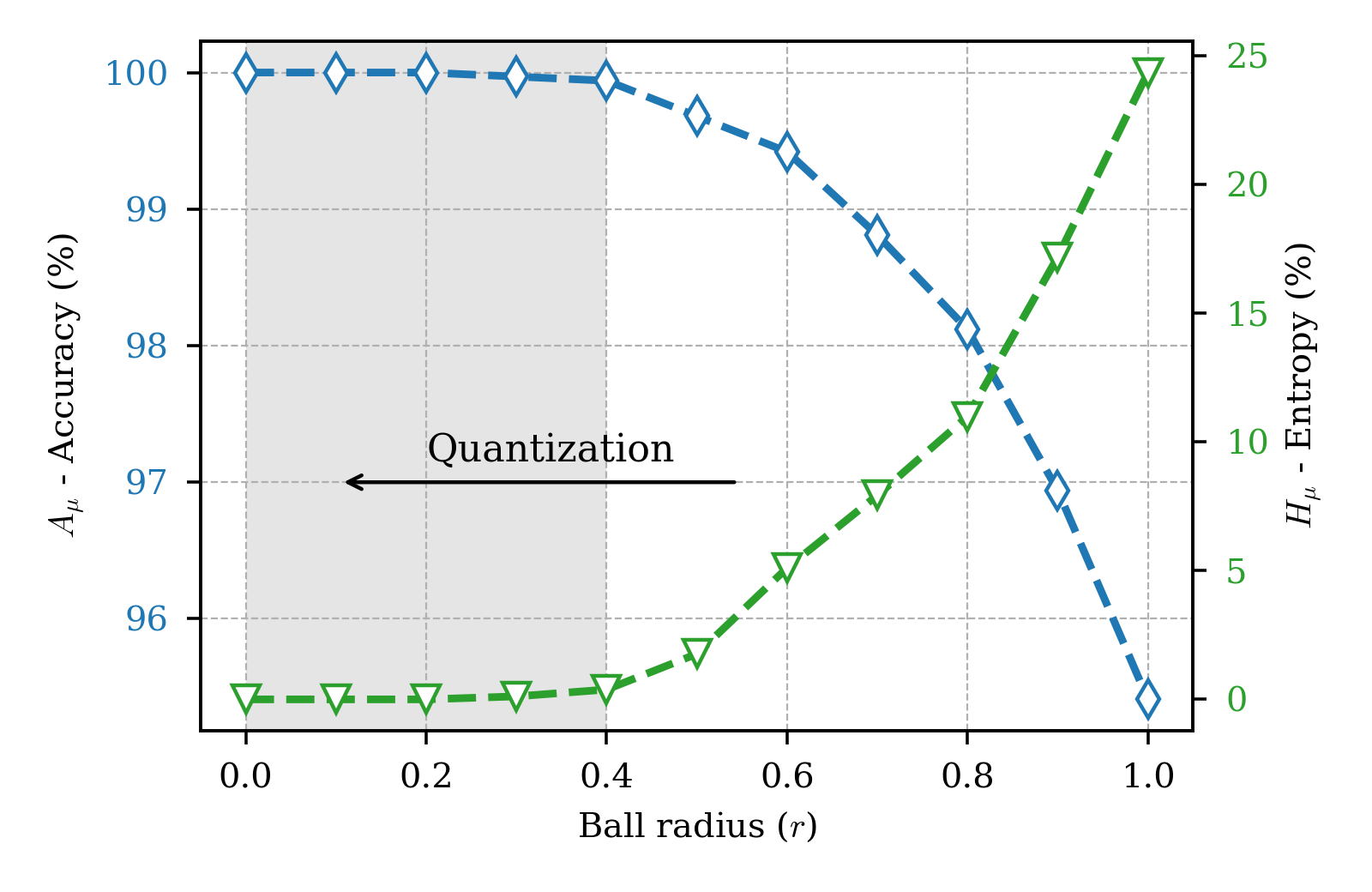}
    \caption{Tradeoff semantic vs. effectiveness.}
    \label{fig:perf_comparison_ball_tradeoff}
\end{figure}

\vspace{0.1cm}
\noindent
\textbf{From semantic to communication effectiveness.} 
In \fig{fig:perf_comparison_ball_tradeoff}, we evaluate the impact of the ball radius $\ballradius$ in Eq. \eqref{eq:new-transport-optimization} on the equalization performance. When $\ballradius$ tends to zero, the accuracy of the communication increases as it becomes less sensitive to syntactic channel noise. Conversely, the entropy of the codebook decreases indicating a loss of diversity in the base semantic representation. Indeed, decreasing $\ballradius$ concentrates the points of the target atoms, resulting in a seamless quantization. Consequently, we lose the information about the specificities (\eg digit font or style) of each image projected into this atom.

\section{Conclusion}
We proposed a method for equalizing semantic channels in multi-user semantic communications, where senders and receivers employ distinct languages. In our proposed solution, language mismatch is first modeled by means of measurable transformations over semantic representation spaces. Then, we proposed a method for learning a codebook of transformations based on optimal transport theory, which models such transformations as transportation maps. Hence, the problem of semantic channel equalization translates to selecting the appropriate transformation in the codebook, which we achieve with a Bayes-optimal selection strategy. Overall, the proposed method is robust against both syntactic and semantic channel noise, improving transmission accuracy with reduced complexity compared to traditional approaches.

\bibliographystyle{IEEEtran}
\bibliography{biblio}
\end{document}